\documentclass[letterpaper, 10 pt, conference]{ieeeconf}  

\IEEEoverridecommandlockouts                              
\usepackage{xcolor}
\usepackage{enumerate}
\usepackage[utf8]{inputenc}
\usepackage[T1]{fontenc}

\usepackage{graphicx}
\graphicspath{ {./figures/} }

\newcommand\litem[1]{\item{\bfseries#1:}}

\title{\LARGE \bf
MCPrioQ: A lock-free algorithm for online sparse markov-chains
}

\author{Jesper Derehag$^{1}$ and Åke AI Johansson$^{2}$
\thanks{$^{1}$J. Derehag is a senior machine learning engineer at Ericsson AB, Lindholmspiren 11, 417 56 Göteborg, Sweden
        {\tt\small jesper.derehag at ericsson.com}}%
\thanks{$^{2}$Å. Johansson is a senior machine learning engineer at Ericsson AB, Lindholmspiren 11, 417 56 Göteborg, Sweden
        {\tt\small ake.ai.johansson at ericsson.com}}%
}

\begin{document}

\maketitle
\thispagestyle{empty}
\pagestyle{empty}

\begin{abstract}
In high performance systems it is sometimes hard to build very large graphs that are efficient both with respect to memory and compute. This paper proposes a data structure called Markov-chain-priority-queue (MCPrioQ), which is a lock-free sparse markov-chain that enables online and continuous learning with time-complexity of $O(1)$ for updates and $O(CDF^{-1}(t))$ inference. MCPrioQ is especially suitable for recommender-systems for lookups of $n$-items in descending probability order. The concurrent updates are achieved using hash-tables and atomic instructions and the lookups are achieved through a novel priority-queue which allows for approximately correct results even during concurrent updates. The approximatly correct and lock-free property is maintained by a read-copy-update scheme, but where the semantics have been slightly updated to allow for swap of elements rather than the traditional pop-insert scheme.
\end{abstract}

\section{INTRODUCTION}
In recommender-systems or collaborative filtering applications it is relatively common to produce recommendations that is based on some cumulative-sum of probabilities. For example, recommend any number of products such that the probability of finding a product that matches a users preferences is above a certain threshold. E.g. a 90\% probability that a matching item exists in the list of recommended items. This is also a useful context in for example telecommunications systems where a users location may be unknown in the network. As such, a cellular network could be considered as a directed graph where the base stations would be nodes and the physical movement of a user through that network are the edges. If the location (node) of a user is unknown, a telecommunications network would then typically perform a paging operation, essentially querying all nodes if the user resides within that nodes area. This could be considered a recommender-system where one wants to recommend a number of nodes such that the system will find the user with some threshold-probability~\cite{mobility_pred}.
In such a system, it would be highly beneficial to be able to construct the graph while simultaneously being able to query it, with inference being optimized for returning a list of items.

\section{MCPrioQ}
\begin{figure}[thpb]
      \centering
      \framebox{\parbox{3in}{\includegraphics[width=\linewidth]{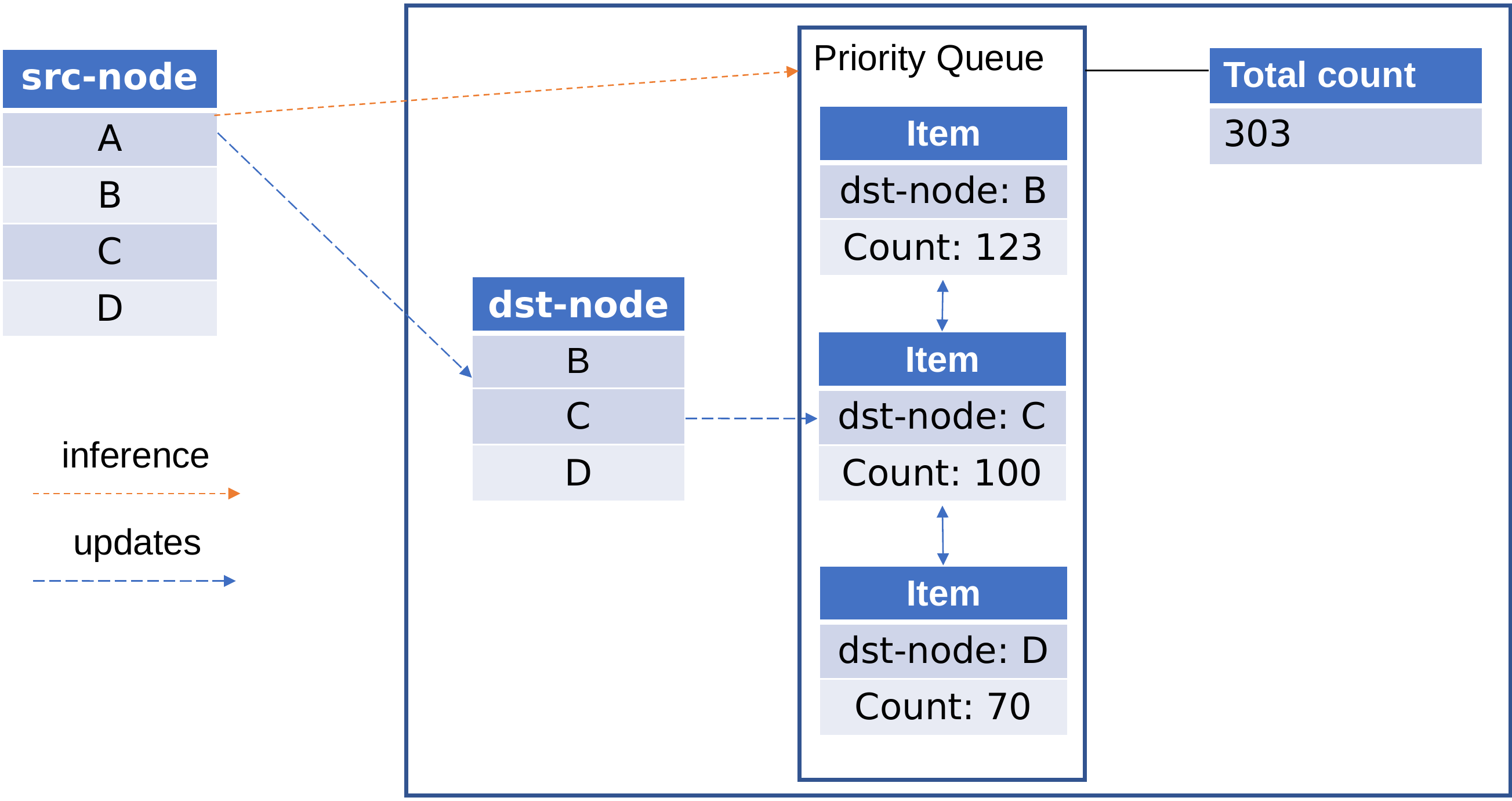}}}
      \caption{Pseudo class-diagram for the data structures}
      \label{fig_class_diagram}
   \end{figure}

The proposed algorithm is a specialized version of a markov-chain where one wants to sort edges based on some counter. The counter could be anything but this implementation is optimized for counters that tend to be incremented in small steps and where increments occur in accordance with the probability of a edge transition. In other words, it is optimized for the assumption that the priority queue is already sorted and any increments will unlikely lead to resorting.

\subsubsection{Node lookup tables}
src-node and dst-node lookup tables need to be lock-free shared data structures. In this particular case we assume these to be Read-copy-update (RCU) hash-tables~\cite{McKenney98}.
Specifically, as we assume that the PriorityQueue is a doubly linked list, which eventually will need elements to be freed, it is important that also the hash-tables are RCU based such that the read-side-critical-sections share the same grace period.

\subsubsection{Priority Queue}
The more interesting part in Fig.~\ref{fig_class_diagram} are the Priority Queue as a sorted doubly linked list, which is central to the contributions of this paper. 
In this particular case we will assume that increments of the priority are small and that the input order is not shuffled. Therefore we argue that the no-inversion case is the normal case for inserts. Furthermore, we know that the search is strictly finding max priority items, so random access is not the normal case. Formally this could be considered a monotone priority queue.
A popular choice for priority queues tends to be a heap which is generally optimized for fast insert and finding the top most important element, while in the cumulative-probability case one wants to find a list of items of cumulative probability. Other popular implementations of priority queues are skip-lists \cite{SUNDELL2005609} which would be a suitable choice for cumulative-probability applications as well. One could argue that we could skip the dst-hash table and only use a skip-list, but since the edge-list is sorted on the edge transition probability and not its node-id, it would mean that the skip-connections cannot be used for any practical benefit. But since we want to order the elements in transition probability and we can assume that updates are not shuffled, average search complexity (for updates) are then equal to the edge probability distribution.
A hash-table with $O(1)$ is hard to beat, but practically the choice may not be that obvious as it would depend on the exact probability distribution and cache-line availability.
For this reason we propose that the priority queue is a doubly-linked list, and the dst-node hash-table is an optional optimization.

\begin{figure}[thpb]
      \centering
      \framebox{\parbox{3in}{\includegraphics[width=\linewidth]{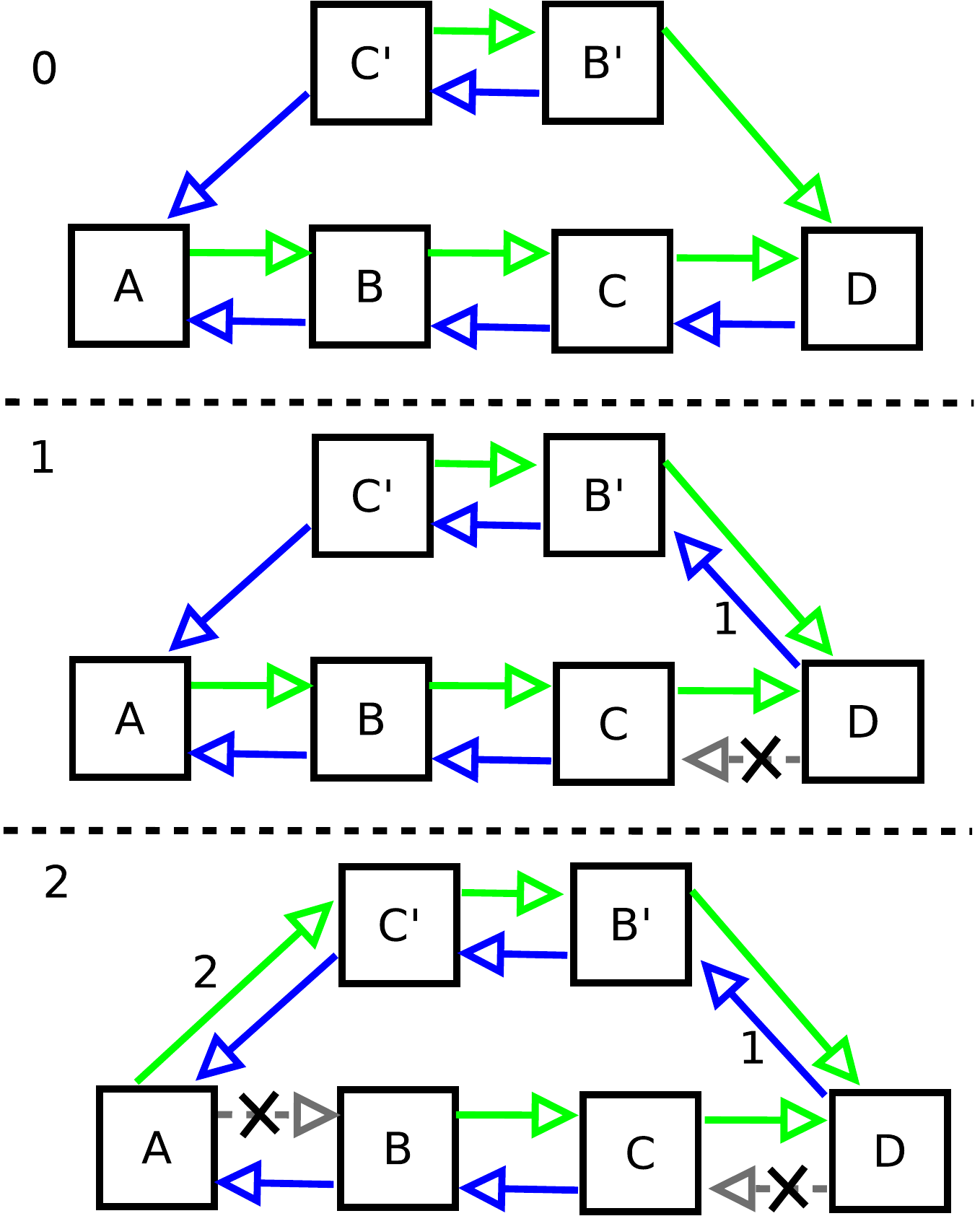}}}
      \caption{Wait-free bubble-sort of doubly linked list}
      \label{fig_dll_swap}
\end{figure}

There are a number of different lock-free implementations of doubly-linked-lists, for ex. \cite{SUNDELL20081008}. But since we are already using an RCU based hash-table we might as well also use an RCU based doubly-linked-list. This will allow us to re-use the read-side-critical-section in the src-node hash-table, and in principle get insert and remove for free. However, the typical RCU doubly-linked-list semantics does not allow for swapping nodes while still maintaining approximately correct order. Specifically one would need to pop-insert elements where concurrent readers could end up with missing nodes in between the pop and insert. Instead we propose an extension to the standard RCU semantics to swap the order of 2 nodes, which allows for an implementation of lock-free bubble-sorting. The swap operation for 2 adjacent nodes are described in Fig. \ref{fig_dll_swap}.

\subsubsection{Counters}
In the case of cumulative transition probability, it is implemented as 2 atomic counters, with one indicating the total number of transitions between two nodes (edge), located as part of the priority queue in Fig.\ref{fig_class_diagram}. There is also an additional counter keeping track of total transitions for all edges given a src-node. The transition probability is then calculated as the edge counter divided by the total number of transitions for the node. By having 2 counters and calculating the probability at inference, there is no need for updating all edges as the probability distribution changes.

\subsection{Updates}
\subsubsection{New edge}
Insert the node(s) in the 2 hash-tables and adding an element at the tail of the priority queue.

\subsubsection{Update of edge} 
Updates of edges would be the normal case, so a $O(1)$ lookup in src-node plus an $O(1)$ lookup in dst-node, and a simple atomic increment in the transition counter inside the priority queue. The normal case would be that they are already in order, so nothing additional would be done. If the update result indicates that the previous node has a smaller count than the current node, those elements should be swapped. Similarly, the previous-previous node needs to be checked for swapping in a bubble-sort style leading to $O(n)$ time-complexity in the worst case. However, since increments are small and assuming non-uniform probability distributions, the normal case would likely be no-swap and in rare cases a single-swap.

\subsection{Inference}
The src-node hash-table of $O(1)$ gives the priority-queue. The priority queue is then the sorted-list of edges with $O(nt)$ for worst-case scenario of a uniform transition probability where $t$ is the probability threshold. Slightly more formally, the complexity is equal to the quantile function of the probability distribution. Practically, oftentimes the edges follow a Zipf distribution, and so intuitively you could say its much more likely that you will only need a small subset of edges early on in the priority queue.

\subsection{Model decay}
One way of dealing with massively large graphs that change over time is to add intentional forgetting, or model decay. As the intended use of this data structure is to be online and which will exist over extended periods of time where we dont want to have to retrain the model at some frequency. One way of dealing with that is to uniformly reduce the transition counts over all edges. For example multiplying all transition counts with $0.5$, either at some threshold over the number of total transitions, or maybe more robustly at some frequency that reflects the probability of graph-topology changes. This will maintain the probability distribution, but as some transition counts reaches $0$, that will indicate that edge is no longer used and as such it can be removed. This will allow the graph to remain roughly up to date, although with some added convergence delay for the probability distributions.

\addtolength{\textheight}{-12cm}   





\bibliographystyle{ieeetr}
\bibliography{refs}

\end{document}